\def\year{2019}\relax
\documentclass[letterpaper]{article} 
\usepackage{aaai19}  
\usepackage{times}  
\usepackage{helvet}  
\usepackage{courier}  
\usepackage{url}  
\usepackage{graphicx}  
\usepackage[pdftex]{hyperref}
\usepackage{amsmath}
\usepackage{color}
\usepackage{textcomp}
\usepackage{booktabs}
\usepackage{ccicons}
\usepackage{todonotes}
\usepackage{amsfonts}
\usepackage{framed}
\usepackage{enumitem}

\makeatletter
  \let\@internalcite\cite
  \def\cite{\def\citeauthoryear##1##2{##1, ##2}\@internalcite}
  \def\shortcite{\def\citeauthoryear##1{##2}\@internalcite}
  \def\@biblabel#1{\def\citeauthoryear##1##2{##1, ##2}[#1]\hfill}
\makeatother

\usepackage[algoruled,lined,linesnumbered,boxed]{algorithm2e}
\SetKwProg{Fn}{Procedure}{}{}

\usepackage{amsthm}
\DeclareMathOperator*{\argmax}{arg\,max}

\newcommand{\cA}{\mathcal{A}}
\newcommand{\cE}{\mathcal{E}}

\newcommand{\cN}{\mathcal{N}}

\newcommand{\tit}[1]{\textit{#1}}

\newcommand{\Dom}{\textit{Dom}}
\newcommand{\Attr}{\textit{Attr}}

\newcommand{\tright}{\triangleright}

\newcommand{\figref}[1]{Figure~\ref{fig:#1}}

\newcommand{\algref}[1]{Algorithm~\ref{alg:#1}}

\begin{document}
%
\title{Learning Optimal and Near-Optimal Lexicographic Preference Lists}
\author{
	Ahmed Moussa\\
	School of Computing\\
	University of North Florida\\
	Jacksonville, FL\\
	nagar@aucegypt.edu
	\And
	Xudong Liu \\
	School of Computing\\
	University of North Florida\\
	Jacksonville, FL\\
	xudong.liu@unf.edu
}
\maketitle

\begin{abstract}
  We consider learning problems of an intuitive and concise preference
  model, called \tit{lexicographic preference lists} (\tit{LP-lists}).
  Given a set of examples that are pairwise ordinal preferences over
  a universe of objects built of attributes of discrete values,
  we want to learn (1) an optimal LP-list that decides the maximum number of
  these examples, or (2) a near-optimal LP-list that decides as many examples
  as it can.
  To this end, we introduce a dynamic programming based algorithm and
  a genetic algorithm for these two learning problems, respectively.
  Furthermore, we empirically demonstrate that the sub-optimal models computed
  by the genetic algorithm very well approximate the de facto optimal
  models computed by our dynamic programming based algorithm, and that the genetic
  algorithm outperforms the baseline greedy heuristic with higher accuracy
  predicting new preferences.
\end{abstract}

\section{Introduction}

Preferences are ubiquitous and important to research fields such as
recommender systems, decision making and social sciences.
In this work, we study preference relations of objects that are
combinations of values in discrete attributes.
Preference relations vary and the research community has seen promising
preference models such as graphical models such as
conditional preference networks \cite{bbdh03},
lexicographic preference trees \cite{booth:learningLP,conf/adt13/LiuT,conf/aaai15/LiuT}, and
lexicographic preference \cite{conf/gcai16/liuT,conf/foiks18/LiuT},
and logical models such as penalty logic \cite{de1994penalty}, 
possibilistic logic \cite{dubois1994possibilistic}, and
answer set optimization \cite{brewka2003answer}.
We focus on a particular lexicographic preference model, called
\tit{lexicographic preference lists}, or \tit{LP-lists} for short,
an intuitive and concise model
describing the importance ordering of the attributes and the preference
orderings of the values in the attributes \cite{fishburn1974exceptional}.
In particular, we examine the learning problem of LP-lists for a given
description of the set of objects and a given set of examples.
This problem is considered in two settings: learning the LP-lists that
agree with the maximum, or close to the maximum, number of examples.

Learning optimal LP-lists with maximum number of satisfied examples has been
proven NP-hard \cite{schmitt2006complexity,conf/aaai15/LiuT}.
In this paper, we introduce a dynamic programming based approach that
learns optimal LP-lists in exponential time in the number of attributes and
the size of attribute domains.
Despite this exponentiality, it is a considerable reduction from the factorial
time complexity of the brute force algorithm that checks every ordering
of attributes and every preference order within each attribute domain.
We show the optimality of our algorithm, and present experimental results
demonstrating its effectiveness for large domains of objects.

Algorithms to learning near-optimal LP-lists have been proposed in the literature,
including the greedy heuristic algorithm \cite{conf/foiks18/LiuT}.
Local search algorithms have been used to learn other preference models,
e.g., tree-structured conditional preference networks \cite{allen2017learning}.
To the best of our knowledge, learning algorithms based on local search techniques
have not been applied to learning LP-lists.
In this paper, we propose to learn near-optimal LP-lists using 
a \tit{genetic algorithm} \cite{mitchell1998introduction},
one of the promising local search algorithms, where LP-lists are straightforward
represented as chromosome strings.
We conduct empirical evaluation of our genetic algorithm and compare it with
the greedy heuristic and our optimal algorithm.
Our results suggest that the genetic algorithm performs very close to
our optimal algorithm and outperforms the greedy heuristic by a margin.

In the following, we first formally define what LP-lists are.
We then present our optimal algorithm using dynamic programming, as
well as our near-optimal genetic algorithm.
We will discuss our experimental results before we conclude
and point to future work.

\section{Lexicographic Preference Lists}
Let us denote by $\cA=\{X_1, \ldots, X_n\}$  a finite set of 
attributes. Each attribute $X_i \in \cA$ has a finite domain $D_i$ of values
such that $D_i=\{x_{i,1},\ldots,x_{i,m_1}\}$.
The universe $U_\cA$ defined by $\cA$ is the Cartesian product of
the attribute domains $D_1 \times \ldots \times D_n$.
We call elements in $U_\cA$ \tit{objects}.
A \tit{lexicographic preference list} (\tit{LP-list}) over $\cA$ is a list of 
attributes in $\cA$, each labeled by a total order over that attribute
domain. Attributes in a LP-list are distinct and a subset of $\cA$.

Let $L=X_{i_1}\tright \ldots \tright X_{i_p}$ be an LP-list over $\cA$, 
and $o$ and $o'$ two objects in $U_\cA$.
We say that $o$ is at least as good as $o'$ in $L$, denoted $o \succeq_L o'$,
if (1) $o(X)=o'(X)$ for all $X\in \cA$, or (2) there is $i_j\in [i_1,i_p]$
such that $o(X_{i_j}) \succ o'(X_{i_j})$ and $o(X_q)=o'(X_q)$ for 
all $q\in [i_1,i_j)$.
Then, we say that $o$ is strictly preferred to $o'$, denoted $o\succ_L o'$, 
if $o\succeq_L o'$ and
$o' \not \succeq_L o$, and that $o$ is equivalent to $o'$, denoted $o\approx_L o'$,
if $o\succeq_L o'$ and $o' \succeq_L o$.

Accordingly, given two objects $o$ and $o$, and an LP-list $L$, 
objects can be compared by an LP-list as follows.
For each attribute $X_{i_j}$ in $L$, starting from the first one,
we check if $o$ has a better (or worse) value on $X_{i_j}$ than $o'$.
If so, we stop and report $o\succ_L o'$ ($o'\succ_L o$, resp.).
Otherwise, $o$ and $o'$ having same value on $X_{i_j}$,
we continue to the next attribute.
If we finish having checked all attributes, we stop and report 
$o\approx_L o'$.
Therefore, this task is done in linear time in the size of the input.

Consider a universe of vehicles of four attributes: 
BodyType ($B$) with values sedan ($s$) and truck ($t$),
Color ($C$) with black ($k$), white ($w$) and blue ($b$),
Make ($M$) with Toyota ($t$) and Chevrolet ($c$),
and 
Price ($P$) with low ($l$), medium ($m$) and high ($h$).
An LP-list can be $B \tright M \tright C$, where $B$ is labeled by
$s\succ t$, $M$ by $t\succ c$, and $C$ by $w\succ b\succ k$.
According to this LP-list, we see that
a medium-priced blue Toyota sedan is preferred to a low-priced white
Chevrolet sedan, and two differently priced black Chevrolet trucks are equivalent.

\section{Dynamic Programming Algorithm}
Given a set $\cE$ of examples and an attribute $X \in \cA$,
we want to compute the optimal local preference ordering of $X$,
denoted $LPO(X)$, that satisfies the maximum number of examples
in $\cE$ just by $X$ alone.

From $\cE$ we first build a matrix $M$ where $M_{i,j}$ denotes the number
of examples in $\cE$ that prefer $i$ to $j$ on attribute $X$.
Examples with same value of $X$ in both are not counted in $M$,
so that $M_{i,j}=0$ if $i=j$.
This first step takes $O(|\Dom(X)|^2)$ both space and time.
Then, we use $M$ to compute the $LPO(X)$ as follows.

Let $S \subseteq \Dom(X)$ be a subset of the domain of $X$,
where $|S|>1$.
We denote by $C(S,\cE)$ the maximum number of examples in $\cE$ that can be
satisfied by any total order on $S$.
Thus, we have the following.
\[
C(S,\cE) = 
  \begin{cases} 
    \max\limits_{i,j\in S}\{M_{i,j}, M_{j,i}\} & \text{if } |S|=2 \\
    \max\limits_{i\in S}\{C(S-\{i\},\cE)+\sum\limits_{j\in S-\{i\}}M_{j,i}\} & \text{if } |S|>2
  \end{cases}
\]
Therefore, $LPO(X)$ is the total order on $X$ that satisfies $C(\Dom(X),\cE)$
examples in $\cE$. Both $LPO(X)$ and $\cE$ are computed using the following
procedure in \algref{alg1}, a dynamic programming based procedure recording 
calculated results in tables $T$ and $L$.

%

\begin{algorithm}[ht]
\KwIn{$\cE$ is the set of example, and $X\in \cA$ is an attribute in $\cA$}
\KwOut{$LPO(X)$}
\Fn{computeLPO($\cE$, $X$):}{
  Create an empty table $L$ s.t. $L[U]$ is an empty list 
    for every $U\subset \Dom(X)$\;
  Create matrix $M$ as described earlier\;
  Create an empty table $T$ s.t. $T[U]=0$ for every $U\subset \Dom(X)$\;

  Set $T[U]$ and $L[U]$ for $|U|=2$ according to above formula\;
  \For{$i\gets 3$ \KwTo $|\Dom(X)|$}{
    \ForEach{$S \subset \Dom(X)$ s.t. $|S|=i$}{
      {\tiny $T[S] \leftarrow \max\limits_{i\in S}\{T(S-\{i\})+\sum\limits_{j\in S-\{i\}}M_{j,i}\}$\;}
      {\tiny $L[S] \leftarrow \argmax\limits_{(L[S-\{i\}],i):i\in S}\{T(S-\{i\})+\sum\limits_{j\in S-\{i\}}M_{j,i}\}$\;}
    }
  }

  \Return $L[\Dom(X)]$\;
}
\caption{\tit{computeLPO}$(\cE,X)$ \ \ \% computes LPO for an attribute for given examples \label{alg:alg1}}
\end{algorithm}

We now analyze the space and time complexity of \algref{alg1} as follows.
The space complexity results from the matrix $M$, and tables $L$ and $T$.
Let $x=|\Dom(X)|$ be the number of values in $X$'s domain, and $m$ the number
of examples in $\cE$.
Then, space complexity is $O((2^x)\cdot \frac{x}{2}+x^2+2^x)=O(x\cdot 2^x)$.
This asymptotic prohibitive space is acceptable, if the size 
of the attribute's domain is relatively small, often the 
case in practice.

To calculate the time complexity, we examine the algorithm closely.
We assume structures $M$, $T$, and $L$ are constant time accessible.
Lines 2 to 5 take time $O(2^x+m+2^x+{x \choose 2})$, respectively.
The loop from line 6 to line 11 considers all subsets $S\subset \Dom(X)$
and $|S|\geq 3$, for each of which $T[S]$ and $L[S]$ are computed.
Each takes time $O(|S|\cdot|S|)$.
So this loop takes time 
$O(\sum\limits_{S\subset \Dom(X) \land |S|\geq 3} |S|^2)$.
Then, we have $\sum\limits_{S\subset \Dom(X) \land |S|\geq 3} |S|^2 =
{x \choose 3}\cdot 3^2 + \ldots + {x \choose x}\cdot x^2 \leq
({x \choose 3}+\ldots+{x \choose x})\cdot x^2 \leq
2^x\cdot x^2$.
Therefore, the time complexity of \algref{alg1} is $O(m+x^2\cdot 2^x)$.
This is a clear reduction from the factorial performance of the 
brute-force approach that checks all permutations of $\Dom(X)$.

Let us denote by $\Attr(T)$ the set of attributes in $\cA$ labeling the nodes in LPL $T$,
by $\alpha|_T$ the partial obtained from object $\alpha$
restricted to attributes showing up in $T$.
Then, we define $\cE|_T=\{(\alpha|_T,\beta|_T): (\alpha,\beta)\in\cE \text{ 
and } \alpha|_T\not= \beta|_T\}$ 
to be the multi-set of examples obtained from $\cE$ restricted to attributes
showing up in $T$.

We say an LPL is \tit{optimal} to $\cE$ if it satisfies the maximum number of examples
in $\cE$.
Inspired by the Held-Karp algorithm \cite{held1962dynamic}, we see that,
if an LPL $T$ is optimal to $\cE$, then every $T$'s subtree $T'$ rooted at $r$ is
optimal to $\cE|_{T'}$. Clearly, this property is true because, were the subtree
$T'$ to be not optimal, $T$ could be changed to satisfy more examples by altering
the order of $\Attr(T')$ in $T'$.

We devise the \algref{alg2} to learn optimal lexicographic preference lists.
It is brute-force enhanced by memorizing the optimal subtrees for all subsets
of $\cA$.

\begin{algorithm}[ht]
\KwIn{$\cE$ is the set of example, and $\cA$ is the set of attributes}
\KwOut{$LPL$}
\Fn{computeLPL($\cE$, $\cA$):}{
  Create an empty table $T$ s.t. $T[U]=0$ for every $U\subset \cA$\;
  Create an empty table $L$ s.t. $L[U]$ is an empty list 
    for every $U\subset \cA$\;
  Set $T[U]$ and $L[U]$ for $|U|=1$ using computeLPO in \algref{alg1}\;
  \For{$i\gets 2$ \KwTo $|\cA|$}{
    \ForEach{$S \subset \cA$ s.t. $|S|=i$}{
      {\tiny $T[S] \leftarrow \max\limits_{X\in S}\{T(S-\{X\})+C(\Dom(X),\{(\alpha,\beta)\in \cE:\alpha(Y)=\beta(Y) \text{ for all } Y\in S-\{X\}\})\}$\;}
      {\tiny $L[S] \leftarrow \argmax\limits_{(L[S-\{X\}],computeLPO(X)):X\in S}\{T(S-\{X\})+C(\Dom(X),\{(\alpha,\beta)\in \cE:\alpha(Y)=\beta(Y) \text{ for all } Y\in S-\{X\}\})\}$\;}
    }
  }

  \Return $L[\cA]$\;
}
\caption{\tit{computeLPL}$(\cE,\cA)$ \ \ \% computes optimal LPL for given examples \label{alg:alg2}}
\end{algorithm}

Let us consider the space and time complexities of \algref{alg2}.
We let $n=|\cA|$ be the number of attributes, and 
$\bar{x}=\max\{|\Dom(X)|: X\in \cA\}$ the maximum attribute domain size.
As with the space complexity, the algorithm uses tables $T$ and $L$, and
the space designated by the calls to \algref{alg1}.
The size of $L$ is bounded by $\sum\limits_{0\leq i\leq n} {n \choose i}
\cdot i \cdot \bar{x} \leq n\cdot \bar{x}\cdot 
\sum\limits_{0\leq i\leq n} {n \choose i} = n\cdot \bar{x}\cdot 2^n$.
This gives us a space complexity of 
$O(2^n+n\cdot \bar{x}\cdot 2^n+\bar{x}\cdot 2^{\bar{x}})$, which is
$O(n\cdot \bar{x}\cdot 2^n+\bar{x}\cdot 2^{\bar{x}})$.

Lines 2 to 4 take time $O(2^n+2^n+n\cdot (m+\bar{x}^2\cdot 2^{\bar{x}}))$ respectively.
The loop from line 5 to line 10 takes time
$\sum\limits_{S\subset \cA \land |S|\geq 2} |S|\cdot (m+\bar{x}^2\cdot 2^{\bar{x}})$.
Therefore, the time complexity of \algref{alg2} is
$O(2^n+2^n+n\cdot (m+\bar{x}^2\cdot 2^{\bar{x}})+(m+\bar{x}^2\cdot 2^{\bar{x}})\cdot 2^n\cdot n)$,
which is $O((m+\bar{x}^2\cdot 2^{\bar{x}})\cdot 2^n\cdot n)$.

\section{Genetic Algorithm}
The idea of the genetic algorithm is inspired from the natural selection 
theory in biology. Generally, the population's fitness will increase until a 
steady state. In such steady state, no improvement can be done once the 
population has reached this stable state. This state could contain a 
global or local optimal solution. 

Each candidate solution or chromosome is composed of many traits, features, 
attributes, or simply ``genes" with each gene being one value 
or ``allele". For our learning problem of LP-lists, chromosomes are encoded 
as follows a string of attributes and values in their domains, 
where upper-case letters represent attributes and lower-case letters
represent values the attributes can be of. 
Taking the previous example of the LP-list in the cars domain:
$B \tright M \tright C$ with the same local preference orderings.
Its chromosome representation clearly is
``Bst Mtc Cwbk".

Using the fitness function that returns the number of correctly classified
examples by the LP-list,
we devise the genetic algorithm as follows.
Step 1: create 100 random LP-list as initial chromosomes.
Step 2: select the top 50 chromosomes overall according to the fitness
function and let them produce two children by crossover and mutation respectively.
Step 2 is repeated for 100 generations before termination.
In step 2, crossover is achieved by shuffling the ordering of
the attributes in the chromosomes, and mutation by shuffling the ordering of
values of a randomly chosen attribute.

\section{Results}
In this section, we present our empirical analysis of our two algorithm:
the dynamic programming based algorithm, for which we call \tit{DPA}, and
the genetic algorithm, which we short-hand to \tit{GA}.

To evaluate our algorithms DPA and GA, we take the greedy algorithm as a
baseline and perform empirical analysis on
sets of examples given by hidden randomly generated LP-lists.
The examples are produced with a noise percentage of examples that are
flipped to create inconsistent examples to simulate practical settings.

Domains of 10 attributes, each of 5 values, are used for our experiments.
Thus, the universe contains $5^{10}$ objects, giving
${5^{10} \choose 2}\approx 5\times 10^{13}$ possible examples.
We first generate a random LP-list of these attributes with random
orderings as their local preferences, and a set $D$ of random examples
for training and testing.
Then, set $D$ is processed based on a noise percentage $\cN$: 
$|D|\cdot \cN$ examples are randomly selected and flipped.
We reserve 80\% of $D$ to train an LP-list model and the other 20\% to
test it.
Our experiments are for $\cN$ of value 15\%, and for 
$D$ of size $10^3, 10^4,\ldots,10^6$.
The instance for every $D$ is repeated 5 times and the average accuracies
and computational time are reported as follows.

\begin{figure}[!ht]
  \centering
    \includegraphics[width=0.42\textwidth]{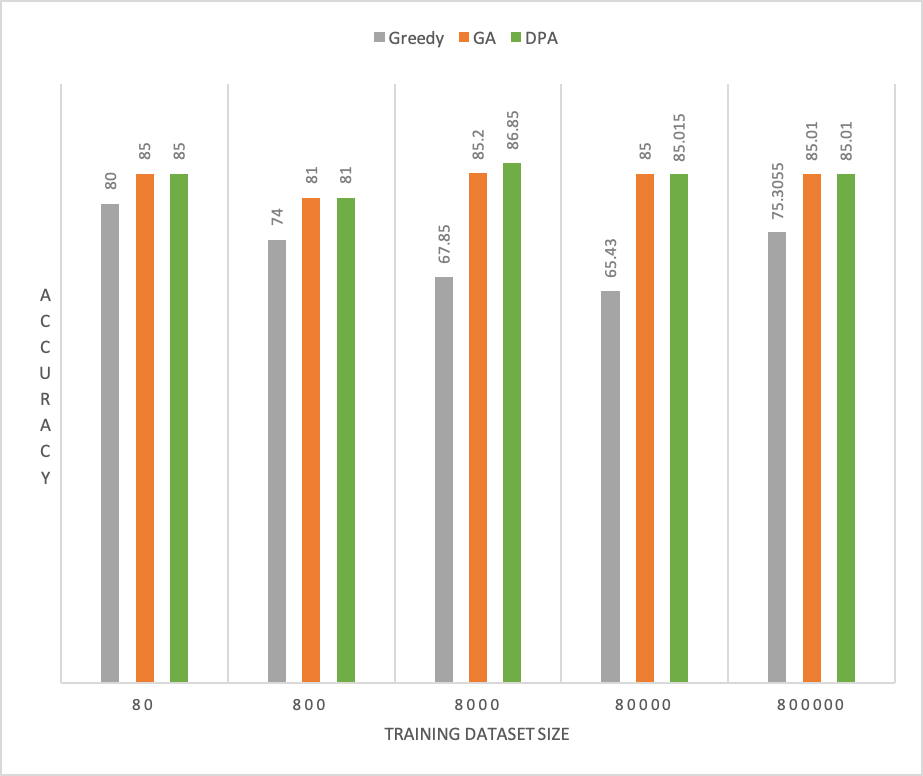}
  \caption{Testing accuracy \label{fig:test}}
\end{figure}

We see, in \figref{test}, that DPA obtains the highest accuracy on the testing
examples.  GA finishes as a very close second, within 1\% compared to DPA.
Greedy finishes last.  We attribute this to the fact that our GA's stochastic
beaming start with multiple LP-lists and generations of improvements of them.

\begin{figure}[!ht]
  \centering
    \includegraphics[width=0.42\textwidth]{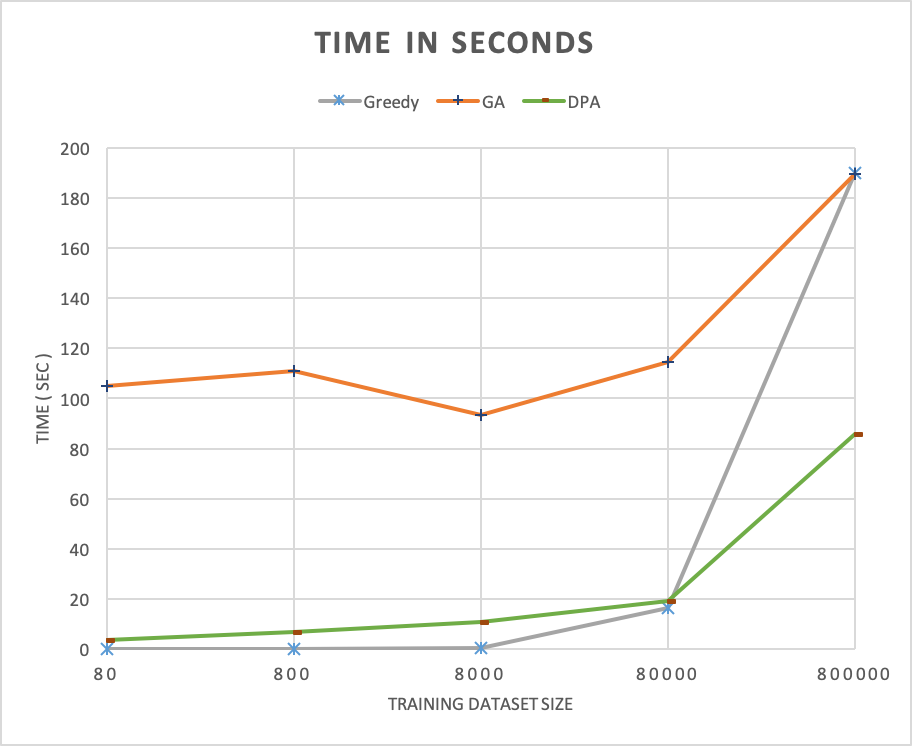}
  \caption{Computational time \label{fig:time}}
\end{figure}

\figref{time} shows the total computational time, including both training and
testing, for various training data sizes.
Clearly, DPA, despite of its exponential time complexity, outperforms
GA on all datasets.
This is because the computational time of GA accumulates over generations.
Greedy takes the least amount of time until the size of the training set
picks up to very large.
This is attributed to the larger constant in the asymptotic notion
of Greedy than that of DPA.
GA takes the most time as it goes through generations of the selecting processes.

\section{Conclusion and Future Work}
We studied the learning problems of LP-lists, a preference formalism
that is intuitive and concise over objects consisting of categorical attributes.
We introduced an algorithm DPA that computes optimal LP-lists that decide the most
number of given examples. DPA is based on dynamic programming, and it reduces
the factorial time complexity of the pure brute force algorithm to exponential,
at a cost of exponential space.
Besides, we introduced a genetic algorithm GA for computing near-optimal
LP-lists that satisfy as many given examples as it can.
To evaluate our algorithms, we conducted substantial experiments showing
that, for large example sets of sizes up to 10 million over the universe of
over 9 million objects, DPA outperforms GA and baseline Greedy in both 
testing accuracy
and computational time, with GA being a very close second in accuracy.
For future work, we plan to perform experimental studies on preferential data 
generated from real-world datasets such as classification and regression 
datasets in the machine learning community.
We also intend to extend our algorithms to allow learning more general
lexicographic preference models \cite{conf/aaai15/LiuT}.

\bibliographystyle{aaai}
\bibliography{refs}

\end{document}